\documentclass{bjmc}   
\usepackage{natbib}
\usepackage{graphicx, float}
\usepackage{todonotes}
\usepackage{subfig}
\usepackage{times}
\usepackage{latexsym}
\usepackage[T1]{fontenc}
\usepackage[utf8]{inputenc}
\usepackage{microtype}

\begin{document}


\title{Similarity of Sentence Representations in Multilingual LMs: Resolving Conflicting Literature and a Case Study of Baltic Languages}

\titlerunning{Similarity of Sentence Representations in Multilingual LMs}  

\author{Maksym DEL, Mark FISHEL}
\authorrunning{Del and Fishel}
\institute{  Institute of Computer Science,
  University of Tartu, Estonia}
\emails{\{maksym,mark\}@tartunlp.ai}


\maketitle      

\begin{abstract} 

Low-resource languages, such as Baltic languages, benefit from Large Multilingual Models (LMs) that possess remarkable cross-lingual transfer performance capabilities. 
This work is an interpretation and analysis study into cross-lingual representations of Multilingual LMs.
Previous works hypothesized that these LMs internally project representations of different languages into a shared cross-lingual space. However, the literature produced contradictory results.
In this paper, we revisit
the prior work claiming that "BERT is not an Interlingua"
and show that different languages do converge to a shared space in such language models with another choice of pooling strategy or similarity index. Then, we perform cross-lingual representational analysis for the two most popular multilingual LMs employing 378 pairwise language comparisons. We discover that while most languages share joint cross-lingual space, some do not. However, we observe that Baltic languages do belong to that shared space.\footnote{The code is available at \url{https://github.com/TartuNLP/xsim}}

\end{abstract}
%
%
\section{Introduction}


Multilingual Language Models such as mBERT \citep{devlin-etal-2019-bert} or XLM-R (XLM-Roberta; \citealt{conneau-etal-2020a-unsupervised}  achieve remarkable results on a variety of cross-lingual transfer tasks  \citep{hu2020extreme, lian2020xglue}. Many works tried to understand how these models represent multiple languages but achieved incomplete and even conflicting results. 

Notably, \citet{muller-etal-2021-first}, \citet{conneau-etal-2020b-emerging} and \citet{singh-etal-2019-bert} performed representational similarity analysis comparing encoded sentences in different languages. However, they come up with two opposite conclusions.

\begin{figure}[b]
    \centering
    \includegraphics[width=\textwidth]{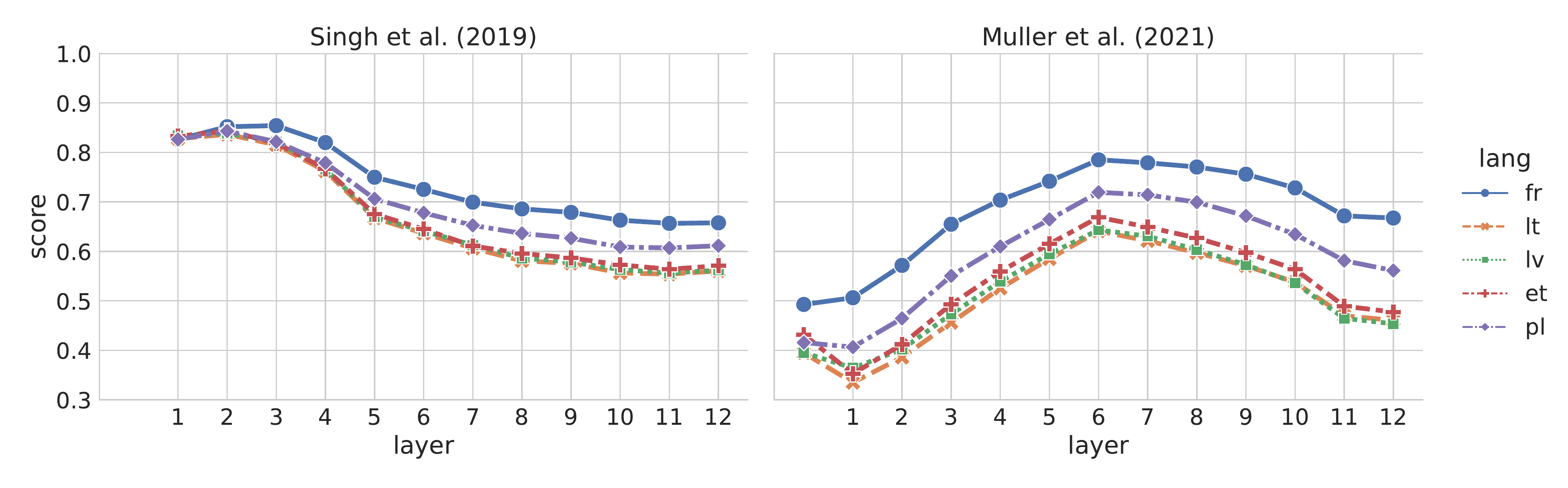}
    \caption{Similarity of mBERT representations decreases (left) vs increases (right). Reproduced from \citet{singh-etal-2019-bert} and \citet{muller-etal-2021-first} on similar data.}
    \label{fig:mbert-cls}
\end{figure}

In particular, \citet{singh-etal-2019-bert} concluded that ``mBERT is not an Interlingua''. They observed that cross-lingual similarity of sentence representations decreased in similarity as the model layer increased. By the term "interlingua," authors most certainly meant the opposite of this pattern.  \citet{muller-etal-2021-first}, on the contrary, found that early layers representations are less aligned than representations from middle layers. See Figure \ref{fig:mbert-cls} for our reproduction of these conflicting results (we choose Estonian, Latvian, and Lithuanian, as leading examples, together with  French and Polish for comparison). In this Figure, we compare languages with English. 

This begs the question of which one we should rely on and why. \citet{muller-etal-2021-first} backs up their representational analysis using probing task in the layer-wise ablation setting; \citet{conneau-etal-2020b-emerging} also directly supports this conclusion. \citet{singh-etal-2019-bert}, on the other hand, provided explanation based on possible tokenization bias issue. In any case, we believe that cross-lingual representational analysis should provide unambiguous results, so we investigate this issue in depth in this work. 

Moreover, we provide a comprehensive study across 378 pairs of languages to have a course-grained view of the cross-lingual similarity. We use Baltic languages as our case study throughout the paper. 

Specifically, we consider the following research questions:
\begin{enumerate}
    \item Which cross-lingual pattern representational similarity suggests in the final analysis? (Answer: convergence pattern from \citet{muller-etal-2021-first}).
    \item Why do results in literature diverge, and how do we interpret them? (Answer: CLS-pooling is a poor choice for sentence summary, and the result is very sensitive to the choice of the similarity index).
    \item What is the recommended way to quantify cross-linguality in multilingual LMs? (Answer: mean-pooling with SVCCA/CKA index aligns with the evidence from the behavior analysis literature better than the other options).
    \item Does the resulting cross-lingual pattern generalize across all languages? (Answer: yes, it does, except for a few outlier languages).
    \item How does representational similarity analysis look for Estonian, Latvian, and Lithuanian? (Answer: they follow the general convergence pattern as the majority of languages and are most similar to each other than to almost all other languages). 
\end{enumerate}


\section{Resolving Conflicting Literature}
In this section, we address the problem of conflicting evidence and conclusions between \citet{singh-etal-2019-bert} and \citet{muller-etal-2021-first} (Figure \ref{fig:mbert-cls}). 

\citet{singh-etal-2019-bert} as a part of their work took a multilingual Bert model\footnote{https://github.com/google-research/bert/blob/master/multilingual.md} and used PWCCA cross-lingual similarity analysis index to measure the discrepancy between languages at each layer. Specifically, they chose a bilingual parallel corpus for a pair of languages (say $lang_A$ and $lang_B$) and passed the pair through the model. Then, they extracted hidden states at each layer, did a CLS pooling, and compared the extracted sentence representations. The resulting trend was that at higher layers, the distance between sentence representations is higher than at the lower layers.

\citet{muller-etal-2021-first} performed a similar procedure with mBERT as part of their work but observed the opposite conclusion: the similarity of language representations is higher at deeper layers and lower at the first layers. In our work, we notice that apart from different datasets, \citet{muller-etal-2021-first} also used a similar different similarity index (CKA) and pooling technique (mean-pooling).

As the conclusion of what happens to cross-lingual representations inside multilingual models is central to both works, we analyze these differences in detail in a shared setting and resolve the conflicting evidence.

\subsection{Setup and Background}

\subsubsection{Data and model}
We do a setup similar to the one of \citet{singh-etal-2019-bert} and use the mBERT-cased model and four parallel datasets (en-et, en-lt, en-lv, en-pl, en-fr; 10k examples for each pair). The parallel corpus is composed of \citet{singh-etal-2019-bert}'s extension of XNLI (Cross-lingual Natural Language Inference) dataset \citep{conneau2018xnli}. We choose Estonian, Latvian, and Lithuanian as our case study and use Polish and French to see how results for Baltic languages compare to high-resource Romance and Slavic languages. 

We embed the source and target sentences with mBERT and pool the \textit{CLS} tokens or perform a mean-pooling over tokens from each layer for each language pair. Next, we compare two parallel sets of sentence representations using the PWCCA, CKA, or SVCCA similarity indexes.

\subsubsection{Representational Similarity Indexes}
\label{sec:metric-bg}

In this subsection, we give a reader brief intuitive explanations and refer to the \citet{cka} for a systematic mathematical description of the correlation similarity indexes. 

All indexes compare two parallel sets of vectors by maximizing correlations (score 1 represents perfect similarity). 

\textbf{CCA} (Canonical Correlation Analysis; \citealt{cca}) is the correlation-based similarity analysis index. As formulated by \citeauthor{pwcca}, CCA "identifies the ’best’ (maximizing correlation) linear relationships (under mutual orthogonality and norm constraints) between two sets of multidimensional variates". CCA score can be a mean of the resulting correlation coefficients.

\textbf{SVCCA} (Singular Vector Canonical Correlation Analysis, \citealt{svcca}) reduces the sensitivity of CCA to particular dimensions by performing Support Vector Decomposition on parallel vectors first and then applying CCA on the resulting components. The number of resulting CCA coefficients is the hyperparameter for the SVCCA, and we use 20 in this work. 

\textbf{CKA} (Centered Kernel Alignment; \citealt{cka}) is another similarity index that works by computing pairwise dot products between two parallel sets of vectors and correlating the resulting distance matrices. 

\textbf{PWCCA} (Projection Weighted Canonical Correlation Analysis; \citealt{pwcca}) is an extension of the original CCA \citep{cca} that weights resulting CCA correlation coefficients based on their importance instead of taking a simple mean. 

Also, we highlight that PWCCA is only invariant to the translation and isotropic scaling. CKA is also invariant to the orthogonal transforms, and SVCCA is invariant to any invertible linear transform.

\subsection{Identifying the Issue}

This section aims to find what caused the discrepancy between results in related works. They use different data and code, but we successfully reproduced the patterns with our datasets. However, they also differ in their choice of pooling strategy for sentence representation (CLS-pooling for \citet{singh-etal-2019-bert} and mean-pooling for \citet{muller-etal-2021-first}) and similarity index (PWCCA for \citet{singh-etal-2019-bert} and CKA for \citet{muller-etal-2021-first}).

\subsubsection{Is this a similarity index issue?}

\begin{figure}[t]
    \centering
    \includegraphics[width=\textwidth]{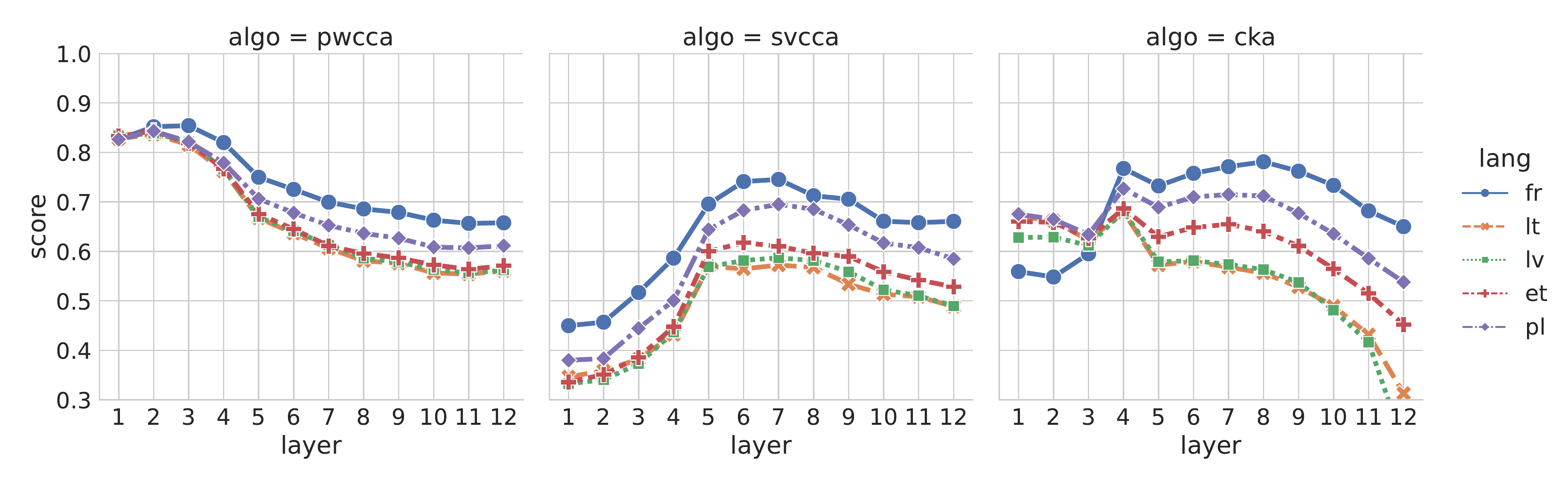}
    \caption{CLS-pooled representations compared using three different similarity measurement algorithms.}
    \label{fig:cls-all_metrics}
\end{figure}

To answer the question of this subsection, we compute all similarities between CLS-pooled representations for all three main indexes. Figure \ref{fig:cls-all_metrics} presents the results. It shows that as we change the similarity measure from PWCCA, we get a convergence pattern similar to one in \citet{muller-etal-2021-first}.

This explains the discrepancy, but the works additionally differ in sentence representation type. In the following subsection, we investigate this issue.

\subsubsection{Is this a pooling strategy issue?}

\begin{figure}[t]
    \centering
    \includegraphics[width=\textwidth]{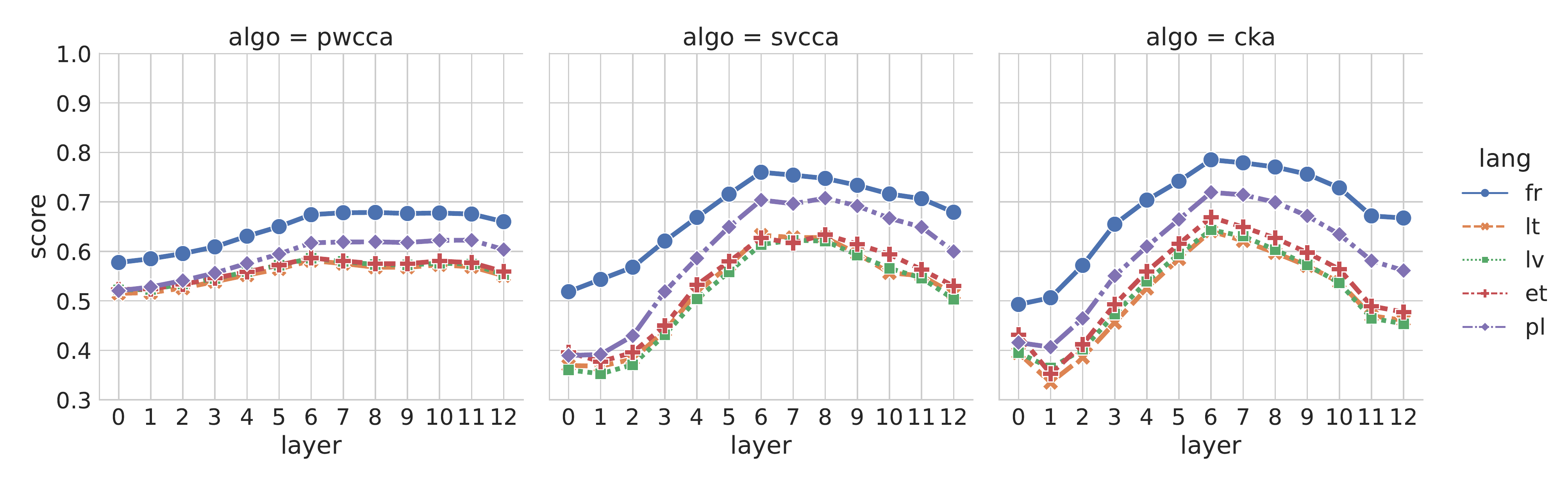}
    \caption{Mean-pooled representations compared using three different similarity measurement algorithms.}
    \label{fig:mean-all_metrics}
\end{figure}

To answer the question of this subsection, we compute all similarity measures for representations, but this time obtained by averaging individual token representations (mean-pooling). Figure \ref{fig:mean-all_metrics} presents the results. It shows that as we change the pooling type from CLS, we get rid of the divergence pattern \citeauthor{singh-etal-2019-bert} observed with CLS-pooling.

PWCCA convergence pattern is less pronounced than other indexes but does not contradict \citet{muller-etal-2021-first}.

The divergence pattern only occurs when we use CLS-pooling and PWCCA metric simultaneously, so in the following subsections, we measure representational similarity with yet another method and question the usefulness of CLS pooling. 

\subsection{Debunking Divergence Pattern}

So we only get a divergence pattern when using PWCCA measure over CLS-pooled representations. In the following subsection, we try cosine similarity as an alternative to the correlational indexes and check the semantic power of CLS pooling on a simple task of cross-lingual sentence matching.

\subsubsection{Cosine similarity}

\begin{figure}[t]
    \centering
    \includegraphics[width=\textwidth]{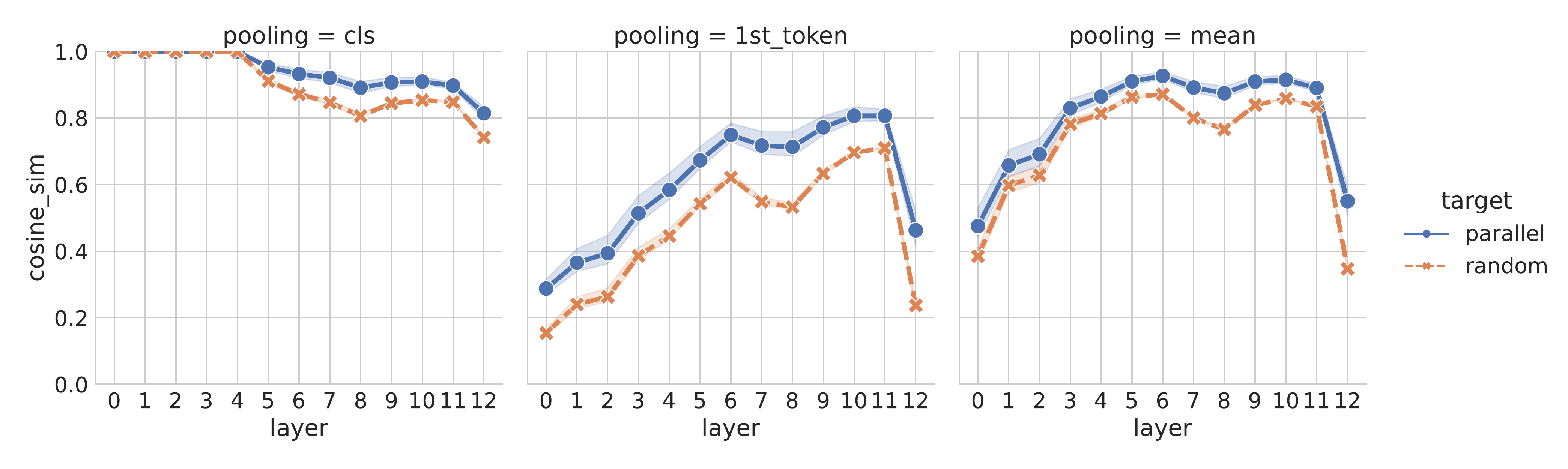}
    \caption{Cosine similarities of sentence representations under three different pooling strategies. "Target" box in legend means that (average) cosine similarity was measured between parallel sentences ("parallel") or arbitrary pairs of sentences ("random").}
    \label{fig:cosine}
\end{figure}

While PWCCA, SVCCA, and CKA are correlation-based indexes, we might also get insight from using simpler alternatives that have proven useful, especially in the NLP domain. One such metric is cosine similarity. We compute the pairwise cosine similarity between English sentences and their target (parallel) translations reporting an average score. We also report the scores for English and sentences and permuted target and compare the similarity over three pooling strategies and report results in Figure \ref{fig:cosine}.

For the first five layers in CLS-pooling, cosine similarities between parallel sentences are close to one, just as cosine for random sentences. So they all point about in the same direction, and there is no straightforward distinction between translations and not translations.  

However, despite representations for CLS being so similar, the question remains how semantically useful they are for cross-lingual similarity analysis and in general. Indeed, even two copies (maybe only slightly altered) of a random matrix would be perfectly similar by all similarity indexes.

\subsubsection{Usefulness of CLS}

Our goal is to find out how cross-lingual mBERT represents languages across layers. A fundamental desired property of cross-linguality is that representations in multiple languages should have close representations. Moreover, these representations should be far closer to representations of other non-parallel sentences.

So let us see how well CLS satisfies these properties. To keep neutral regarding similarity measures, we step away from direct representational similarity analysis and set up a probing task that measures desired properties. We use the same data, and for each English sentence, we find the closest target sentence in the opposite language (out of all 10k targets) and declare "1" if the closest sentence was an actual translation of the source. We declare "0" otherwise. Then we compute the accuracy of this matching task for our language pairs.   

\begin{figure}
    \centering
    \includegraphics[width=\textwidth]{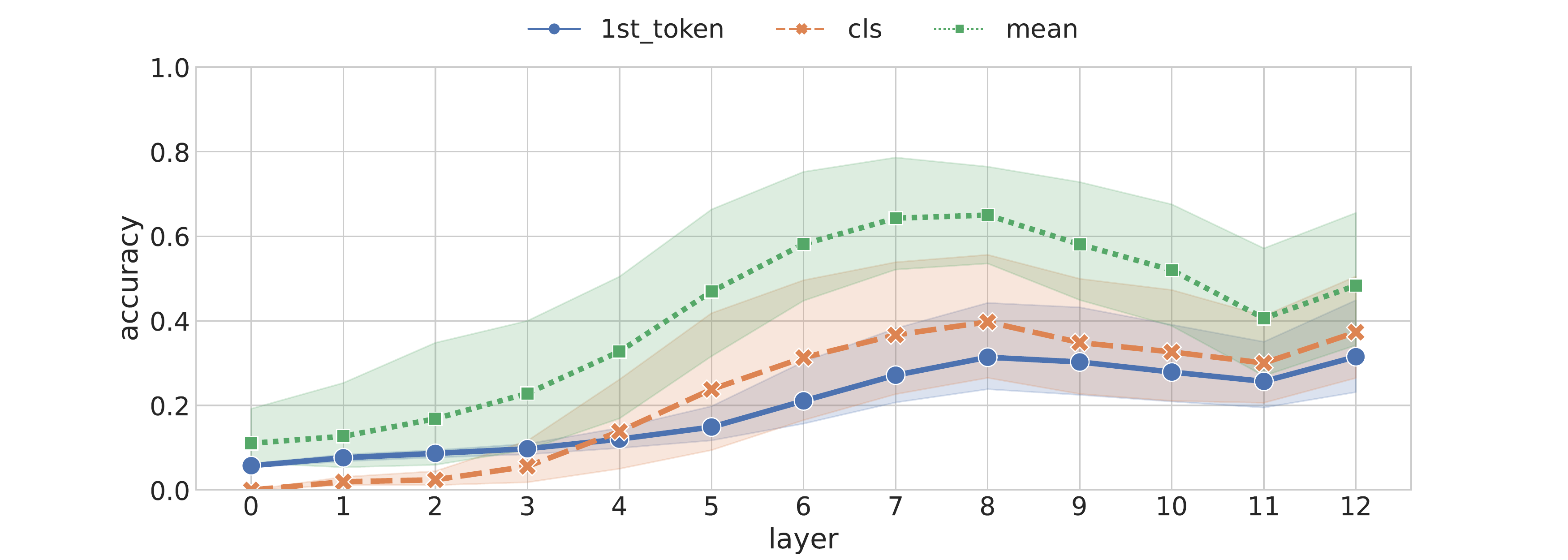}
    \caption{Accuracy of closest sentence vector to each source sentence being an actual translation of this sentence. Measured over three pooling strategies and averaged over four languages (as before).}
    \label{fig:task_all_together}
\end{figure}

We repeat this experiment where we pool the 1st token from each sentence to get a reference point for comparison. We present (averaged over languages) results in Figure \ref{fig:task_all_together}.

The figure shows that accuracy for the CLS-pooling is almost zero at layers 0-4, which suggests it is not a helpful representation to rely upon when measuring cross-linguality, including using CCA-like measures. Matching by the first token is better than matching by CLS at these layers. In later 
layers CLS gets higher than first token pooling, but mean-pooling is about 0.3 accuracy points above. 

While we showed that mean-pooling empirically performs better than CLS-pooling, there remains the question of why it is the case. We explain that CLS embedding does not have a robust explicit signal (representing a sentence meaning) at the layers other than last.

CLS-pooling is used as a sentence representation because, at the last layer, mBERT uses it to predict the next sentence in the corpus during pretraining. The next sentence prediction task uses the CLS token at the last layer. However, CLS at other hidden layers does not have this strong signal to represent the sentence due to the token mixing procedure in the Transformer layers. Each token position carries pieces of information about itself as well pieces about other tokens, and CLS is only slightly more than a regular token at these layers (\ref{fig:task_all_together}. Mean pooling, on the contrary, gathers distributed information about all tokens at each layer and is thus free of the abovementioned issue. 

This section showed that using CLS-pooling and PWCCA measures are suboptimal for cross-linguality representational analysis. They result in a pattern opposite to the expedient one when used together.

\section{Analyzing Language Representations}
In the previous section, we identified that the combination of PWCCA and CLS-pooling is not very suited for cross-lingual analysis. So in this so we use the combination of mean-pooling and CKA. We perform cross-lingual measurements over mean-pooled representations across two pretrained models and 378 pairwise language comparisons.

\subsection{Quantifying cross-linguality Across Languages}
\label{sec:languages}

\begin{figure}%
    \centering
    \subfloat[\centering mBERT]{    \includegraphics[width=0.46\textwidth]{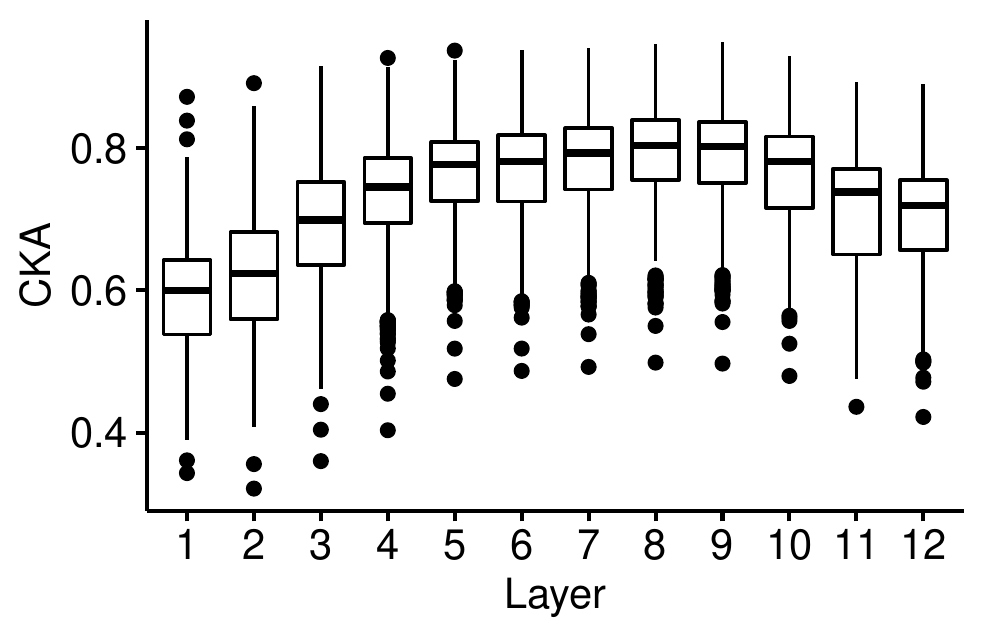}}
    \qquad
    \subfloat[\centering XLM-R]{{    \includegraphics[width=0.46\textwidth]{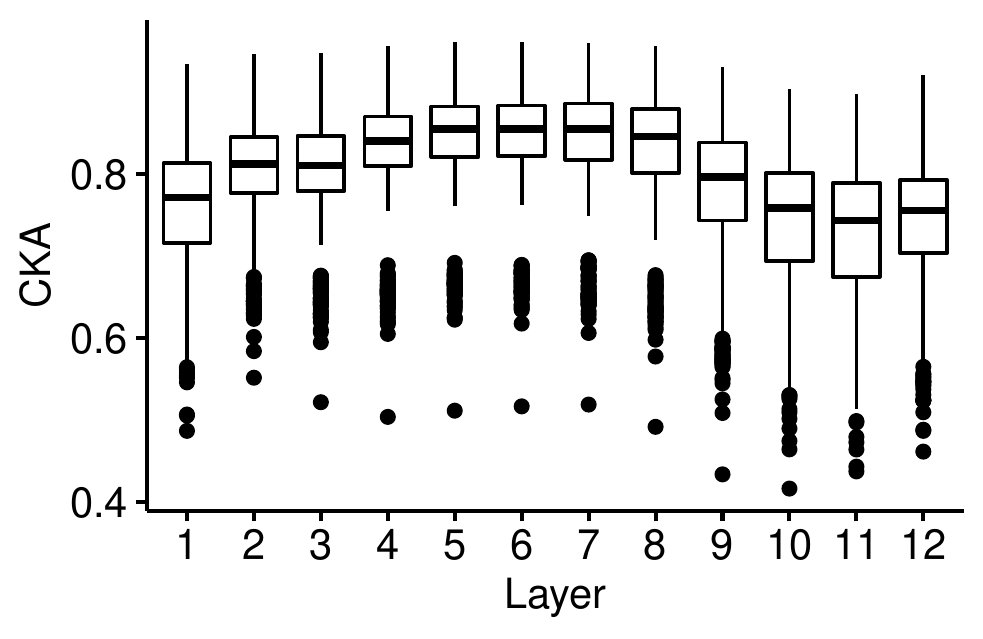}
    }}%
    \caption{Per-layer CKA similarity for mBERT (a) and XLM-R (b) at each layer for combinations of 378 pairwise language comparisons. Dots at the bottom show that some language pairs drastically differ in similarity compared to the vast majority of other language pairs.}%
    \label{fig:interlingua-boxplot}

    \subfloat[\centering mBERT]{    \includegraphics[width=0.46\textwidth]{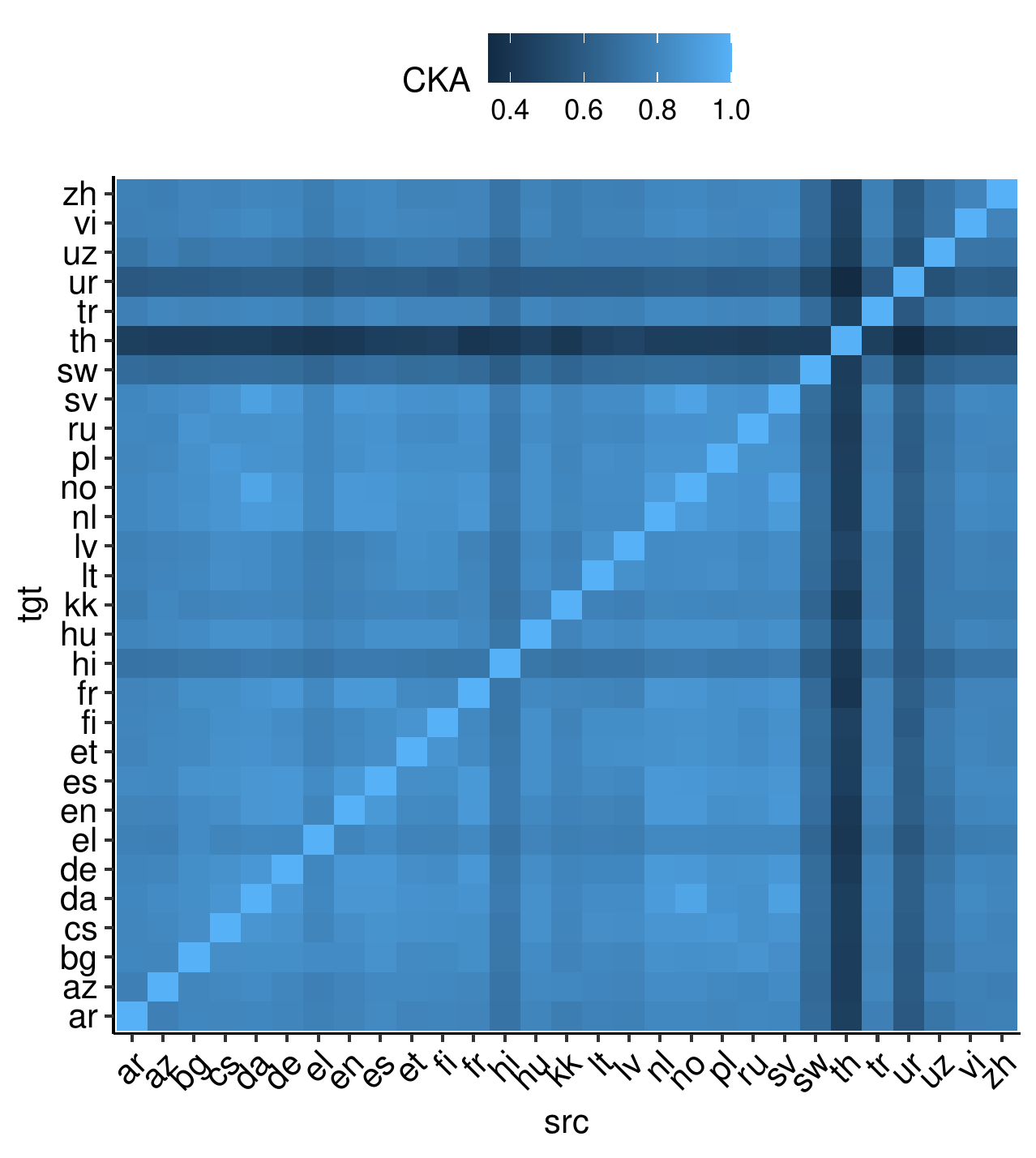}}
    \qquad
    \subfloat[\centering XLM-R]{{        \includegraphics[width=0.46\textwidth]{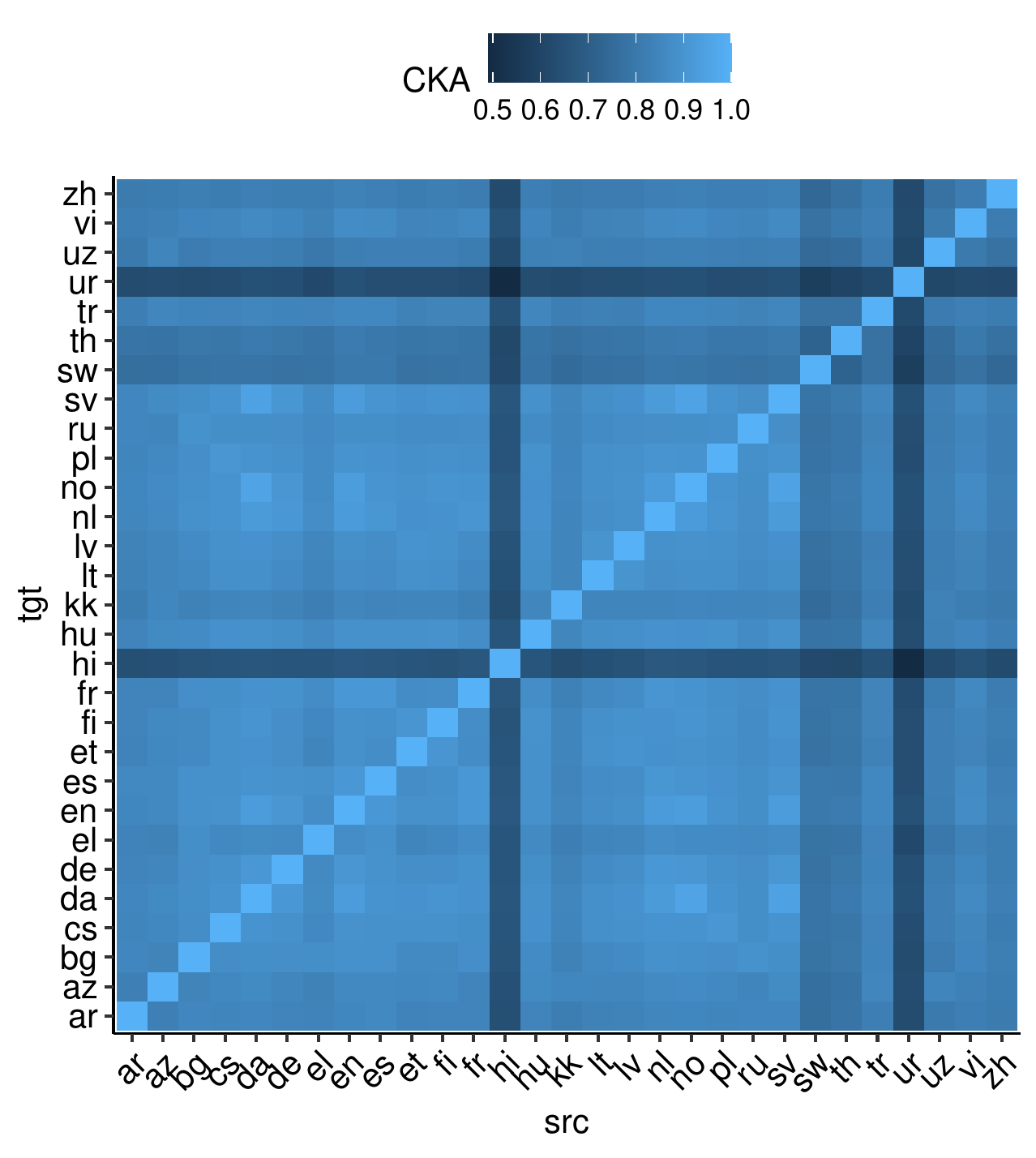}
    }}%
    \caption{All pairwise language similarities at eighth mBERT layer (a) and 7th XLM-R layer (b). There are few languages (Urdu and Hindi and Swahili and Thai) that models seem to locate far away from all others. Estonian, Latvian and Lithuanian do not resemble this property and belong to the main set. }%
    \label{fig:interlingua-heatmap}

    \subfloat[\centering mBERT]{    \includegraphics[width=0.46\textwidth]{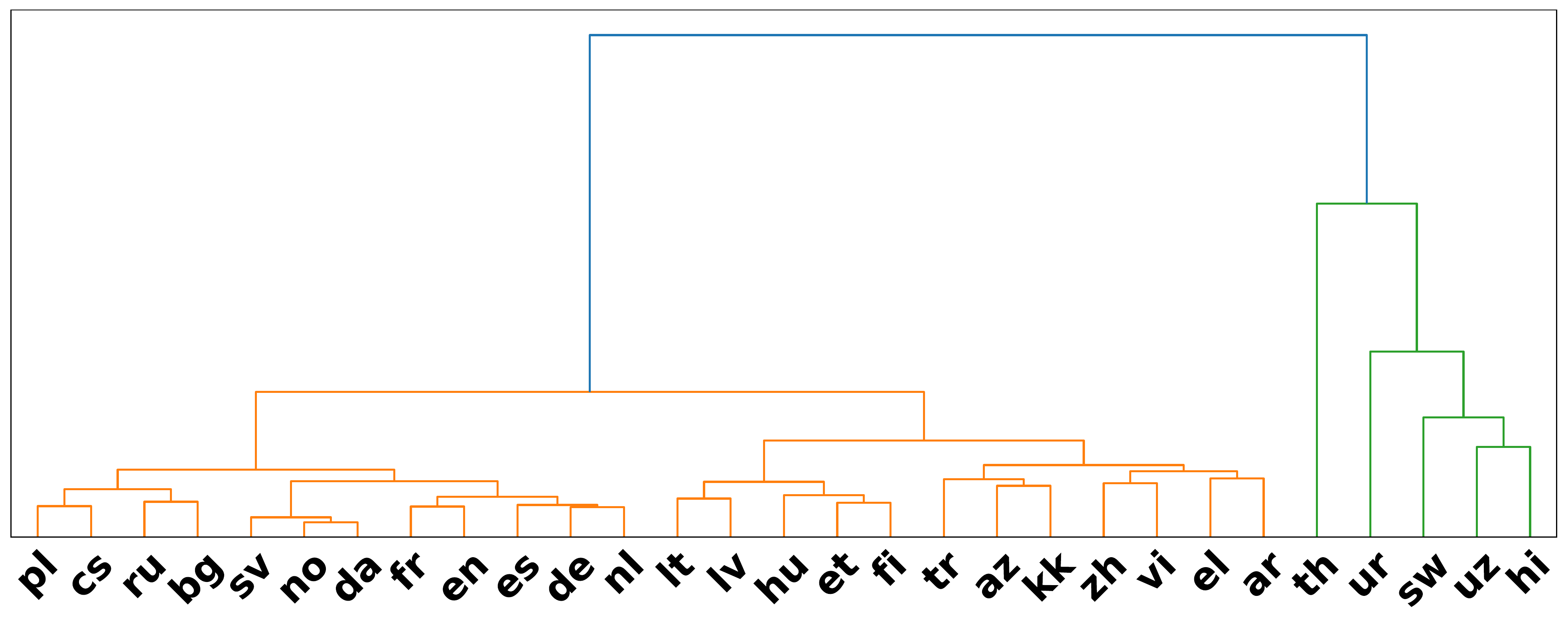}}    \qquad
    \subfloat[\centering XLM-R]{{            \includegraphics[width=0.46\textwidth]{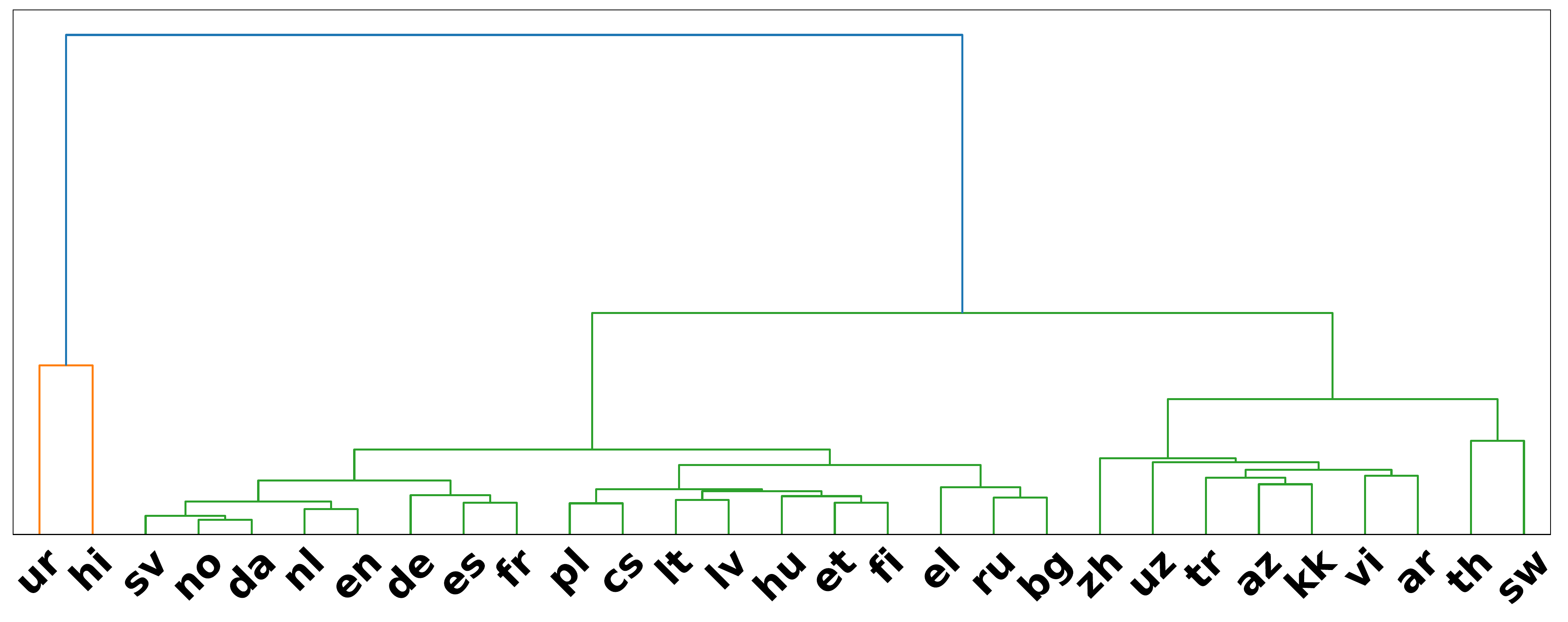}
    }}%
    \caption{Agglomerative clustering for the eighth layer of mBERT (a) and the seventh layer of XLM-R (b) based on CKA distances. We can see that some languages (green for mBERT and orange for XLM-R) branch out from the main branch, and models seem to exclude them from the representational space. Estonian is grouped with Finnish (and then Hungarian), forming the Finno-Ugric branch, while Latvian groups with Lithuanian, forming the Baltic branch. Then these two branches merge into a single "Balto-Finno-Ugric" branch, which suggests these languages can be effectively considered collectively in NLP applications and research.}%
    \label{fig:interlingua-tree}
\end{figure}

\subsubsection{Setup}
This section takes a more in-depth look at cross-linguality in multilingual LMs and explores its structure.

We use all 378 pairwise language comparisons from the extended XNLI dataset \citep{singh-etal-2019-bert}. By computing CKA across all language pairs at all layers, we determine that the seventh layer of XLM-R is the most ``cross-lingual''. We choose to focus on XLM-R in addition to the mBERT as this model is more modern and has shown to be superior to mBERT.

Figure \ref{fig:interlingua-boxplot} also highlights Thai as the most distinct language. \textit{mBERT uncased} pretraining did not include this language. mBERT has more "Excluded" languages, which is logical since it is a weaker cross-lingual model than the XLM-R.

\subsubsection{Results and Discussion}

Figure \ref{fig:interlingua-boxplot} shows the boxplot covering CKA distances between all pairs of languages for XLM-R and mBERT.

Dots at the bottom suggest that there are clear outlier language pairs. The most cross-lingual layer for XLM-R seems to be 7th, so let us open the box at this layer with Figure \ref{fig:interlingua-heatmap}.

The figure suggests that the outliers are due to Urdu and Hindi (possibly also Swahili and Thai).

However, even within the shared cross-lingual space, languages also exhibit certain relationships. Thus we perform an agglomerative clustering on CKA distances to investigate this phenomenon and present the result in Figures \ref{fig:interlingua-tree}.

The linguistic tree in Figure \ref{fig:interlingua-tree} clearly shows that languages in the majority branch (that we consider to be the shared cross-lingual space) structure in a meaningful way. Slavic languages are together, Swahili and Thai are isolated, while Scandinavian languages are again nearby. Urdu and Hindi expectedly occupy their separate branch as outliers. 

The figure also shows that only a small subset of mostly low-resource languages like Swahili and Urdu gets its separate top-level branch (we leave finding the exact criteria for exclusion from the shared space for future work), while other languages, including ones analyzed in \citet{singh-etal-2019-bert}, are a part of the shared cross-lingual space.     

\begin{figure}
\begin{center}
\includegraphics[width=\textwidth]{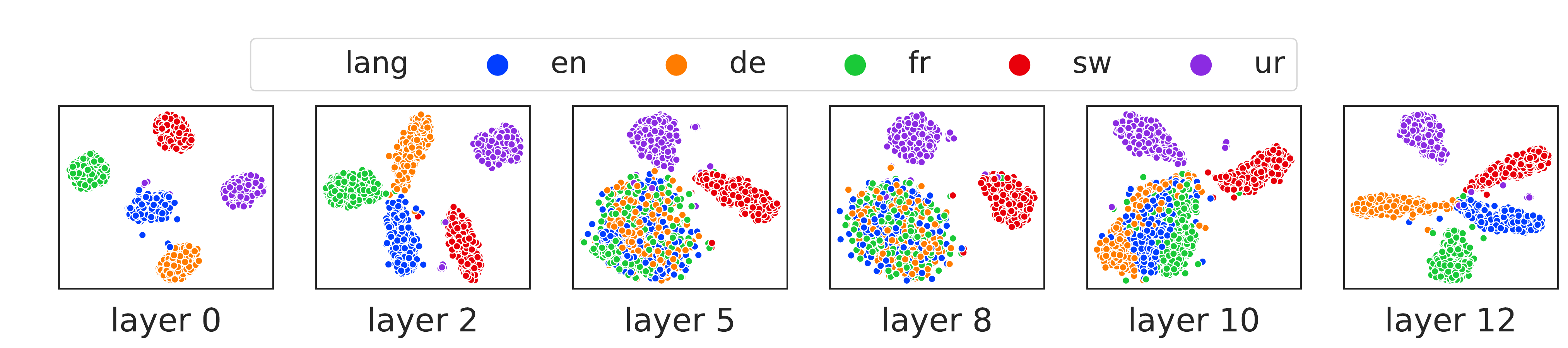}
\end{center}
\caption{Exploration of the cross-linguality pattern via t-SNE. In the first layers (0-2), the representations for all five languages are separated, followed by the middle layers (5-8), where the algorithm could not distinguish between German, French, and English while leaving Swahili and Urdu separated. In the last layers, however, the European selection of languages becomes separated again, finishing with its initial state at the last layers. }
\label{fig:tsne}
\end{figure}

Let us also employ the t-SNE method to demonstrate how models separate extremely low-resource languages (Swahili and Urdu) from the joint space of European Languages (English, German, and French). See Figure \ref{fig:tsne} for the resulting graph.

Due to the t-SNE algorithm's nature, the representational subspaces' shapes and locations do not carry useful information. However, the graph supports our main point that not all languages are a part of joint cross-lingual space. Baltic languages, however, do belong to the joint space, as we showed in Figure \ref{fig:interlingua-tree}.

In summary, in this chapter, we presented a bird's-eye view of how multilingual LMs represent languages across layers. By performing  378 pairwise comparisons we identified that the vast majority of languages share the common interlingual space at the middle layers of the network.

\section{Conclusion}
This paper identified, analyzed, and resolved conflicting literature and derived that mean-pooling with SVCCA/CKA similarity measure is the most suitable choice for the cross-lingual representational similarity analysis. Next, we showed that the pattern is not specific to mBERT and is present in other multilingual language models. Finally, we analyzed 378 pairwise language comparisons and found that not all languages share the cross-lingual space equally. We found, however, that Estonian and Baltic languages are grouped and constitute a part of this shared cross-lingual space.




%
\bibliography{anthology,custom,legacy}

@inproceedings{conneau-etal-2020a-unsupervised,
    title = "Unsupervised Cross-lingual Representation Learning at Scale",
    author = "Conneau, Alexis  and
      Khandelwal, Kartikay  and
      Goyal, Naman  and
      Chaudhary, Vishrav  and
      Wenzek, Guillaume  and
      Guzm{\'a}n, Francisco  and
      Grave, Edouard  and
      Ott, Myle  and
      Zettlemoyer, Luke  and
      Stoyanov, Veselin",
    booktitle = "Proceedings of the 58th Annual Meeting of the Association for Computational Linguistics",
    month = jul,
    year = "2020a",
    address = "Online",
    publisher = "Association for Computational Linguistics",
    url = "https://www.aclweb.org/anthology/2020.acl-main.747",
    doi = "10.18653/v1/2020.acl-main.747",
    pages = "8440--8451",
}

@inproceedings{conneau-etal-2020b-emerging,
    title = "Emerging Cross-lingual Structure in Pretrained Language Models",
    author = "Conneau, Alexis  and
      Wu, Shijie  and
      Li, Haoran  and
      Zettlemoyer, Luke  and
      Stoyanov, Veselin",
    booktitle = "Proceedings of the 58th Annual Meeting of the Association for Computational Linguistics",
    month = jul,
    year = "2020b",
    address = "Online",
    publisher = "Association for Computational Linguistics",
    url = "https://www.aclweb.org/anthology/2020.acl-main.536",
    doi = "10.18653/v1/2020.acl-main.536",
    pages = "6022--6034",
}

@article{cka,
  title={Similarity of Neural Network Representations Revisited},
  author={Simon Kornblith and Mohammad Norouzi and H. Lee and Geoffrey E. Hinton},
  journal={ArXiv},
  year={2019},
  volume={abs/1905.00414}
}

@inproceedings{pwcca,
  title={Insights on representational similarity in neural networks with canonical correlation},
  author={Ari S. Morcos and M. Raghu and S. Bengio},
  booktitle={NeurIPS},
  year={2018}
}

@inproceedings{svcca,
  title={SVCCA: Singular Vector Canonical Correlation Analysis for Deep Learning Dynamics and Interpretability},
  author={M. Raghu and J. Gilmer and J. Yosinski and Jascha Sohl-Dickstein},
  booktitle={NIPS},
  year={2017}
}

@article{cca,
  title={Canonical Correlation Analysis: An Overview with Application to Learning Methods},
  author={D. Hardoon and S. Szedm{\'a}k and J. Shawe-Taylor},
  journal={Neural Computation},
  year={2004},
  volume={16},
  pages={2639-2664}
}

@article{hu2020extreme,
  author    = {Junjie Hu and Sebastian Ruder and Aditya Siddhant and Graham Neubig and Orhan Firat and Melvin Johnson},
  title     = {{XTREME:} {A} Massively Multilingual Multi-task Benchmark for Evaluating
               Cross-lingual Generalization},
  journal   = {CoRR},
  volume    = {abs/2003.11080},
  year      = {2020},
  url       = {https://arxiv.org/abs/2003.11080},
  archivePrefix = {arXiv},
  eprint    = {2003.11080},
}

@article{lian2020xglue,
  author    = {Yaobo Liang and
                 Nan Duan and
                 Yeyun Gong and
                 Ning Wu and
                 Fenfei Guo and
                 Weizhen Qi and
                 Ming Gong and
                 Linjun Shou and
                 Daxin Jiang and
                 Guihong Cao and
                 Xiaodong Fan and
                 Bruce Zhang and
                 Rahul Agrawal and
                 Edward Cui and
                 Sining Wei and
                 Taroon Bharti and
                 Ying Qiao and
                 Jiun{-}Hung Chen and
                 Winnie Wu and
                 Shuguang Liu and
                 Fan Yang and
                 Rangan Majumder and
                 Ming Zhou},
  title     = {{XGLUE:} {A} New Benchmark Dataset for Cross-lingual Pre-training,
                 Understanding and Generation},
  journal   = {CoRR},
  volume    = {abs/2004.01401},
  year      = {2020},
  url       = {https://arxiv.org/abs/2004.01401},
  archivePrefix = {arXiv},
  eprint    = {2004.01401},
}

@inproceedings{Conneau2018xnli,
    title = "{XNLI}: Evaluating Cross-lingual Sentence Representations",
    author = "Conneau, Alexis  and
      Rinott, Ruty  and
      Lample, Guillaume  and
      Williams, Adina  and
      Bowman, Samuel  and
      Schwenk, Holger  and
      Stoyanov, Veselin",
    booktitle = "Proceedings of EMNLP 2018",
    year = "2018",
    pages = "2475--2485",
}

\end{document}